%% file: main.tex
\documentclass[10pt,twocolumn,letterpaper]{article}

\usepackage{wacv}
\usepackage{times}
\usepackage{epsfig}
\usepackage{graphicx}
\usepackage{amsmath}
\usepackage{amssymb}
\usepackage{bbm}
\usepackage{multirow}
\usepackage{stackengine}

\DeclareMathOperator*{\argmax}{arg\,max}
\DeclareMathOperator*{\argmin}{arg\,min}

\wacvfinalcopy 

\def\wacvPaperID{166} 

\ifwacvfinal\fi
\begin{document}

\title{Give me a hint! Navigating Image Databases using Human-in-the-loop Feedback}

\author{Bryan A. Plummer\footnotemark[3] \hspace{1cm} M. Hadi Kiapour\footnotemark[2] \hspace{1cm} Shuai Zheng\footnotemark[2] \hspace{1cm} Robinson Piramuthu\footnotemark[2]\\
\footnotemark[3] Boston University\hspace{10mm} \footnotemark[2] eBay Inc.\\
{\tt\small bplum@bu.edu} \hspace{1cm} {\tt\small  \{mkiapour,shuzheng,rpiramuthu\}@ebay.com}
}

\maketitle
\ifwacvfinal\thispagestyle{empty}\fi

\input{abstract}

\input{introduction}

\input{attribute_embeddings}

\input{active_image_search}

\input{experiments}

\input{scene_experiments}

\input{conclusion}

{\small
\bibliographystyle{ieee}
\bibliography{egbib}
}

\end{document}

%% file: abstract.tex
\begin{abstract}
In this paper, we introduce an attribute-based interactive image search which can leverage human-in-the-loop feedback to iteratively refine image search results.  We study active image search where human feedback is solicited exclusively in visual form, without using relative attribute annotations used by prior work which are not typically found in many datasets. In order to optimize the image selection strategy, a deep reinforcement model~\cite{mnih-atari-2013} is trained to learn what images are informative rather than rely on hand-crafted measures typically leveraged in prior work.  Additionally, we extend the recently introduced Conditional Similarity Network~\cite{veitCVPR2017} to incorporate global similarity in training visual embeddings, which results in more natural transitions as the user explores the learned similarity embeddings.  Our experiments demonstrate the effectiveness of our approach, producing compelling results on both active image search and image attribute representation tasks.
\end{abstract}

%% file: introduction.tex
\section{Introduction}
\begin{figure}[t]
\centering
\topinset{\bfseries Examples of User Feedback}{\topinset{\bfseries(a)}{\topinset{\bfseries(b)}{\includegraphics[width=0.48\textwidth,trim=0cm 3.5cm .0cm 0cm, clip]{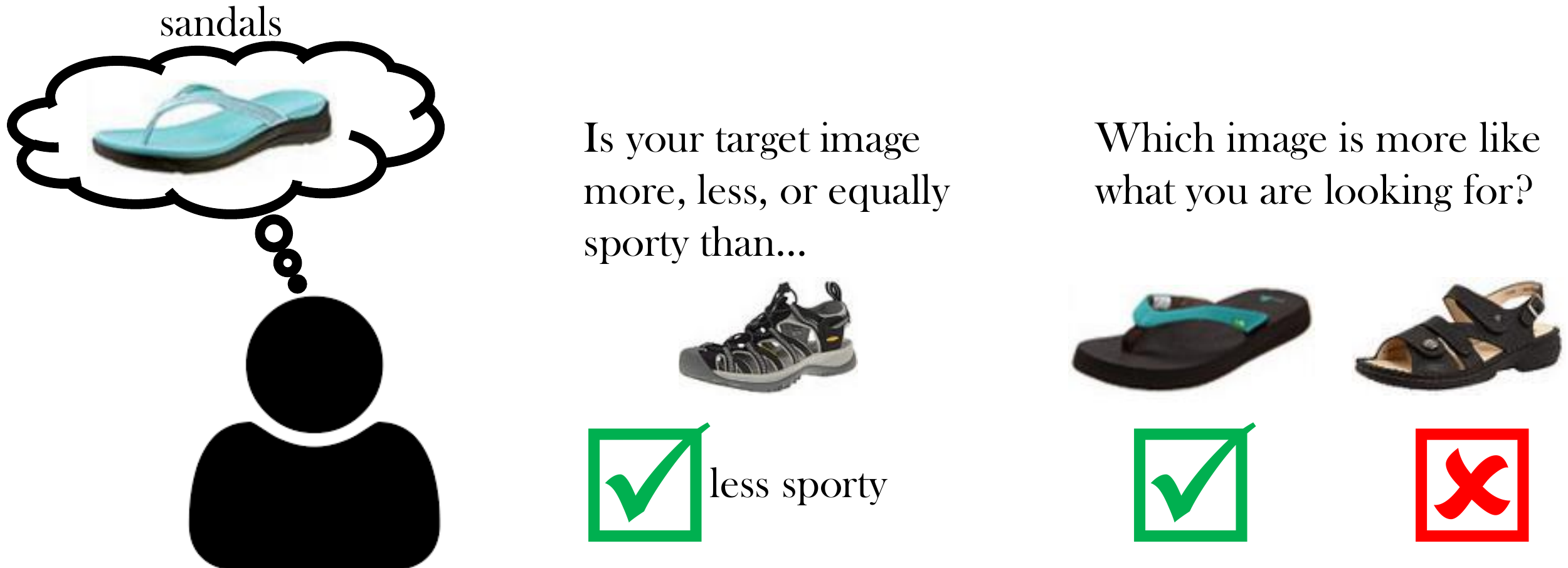}}{.23in}{.55in}}{.23in}{-.53in}}{-.0in}{.21in}
\caption{{\bf Comparison of active feedback types for image search.} Recent work in iterative image search leverages relative attribute annotations enabling the collection of targeted feedback as seen in {\bf (a)}.  However, these relative attributes are not typically collected natively in many datasets.  Instead, we use the meta-data or categorical labels already present in most datasets to build a representation for our active image search approach.  While this relies on more ambiguous feedback since users can define their own similarity criterion, as shown in {\bf (b)}, our experiments show it is sufficient most of the time.}
\label{fig:feedback}
\end{figure}

In image search applications the user often has the mental picture of their desired content. The ultimate goal of image search is to convey this mental picture to the system and overcome the difference between the lower-level image representation and the higher-level conceptual content. Describing the desired image may be time consuming, however, and an image search system may not find the image even with an accurate description.  To remedy this issue, interactive search techniques (\eg ~\cite{caoACMM2010,coxTIP2000,ferecatuICCV2007,FogartyCHI2008,liICME2001,macarthur2000,rasiwaslaTM2007,tongACMMM2001,ZhangACMM2012,Zhao_2017_CVPR,zhouACMMS2003}) obtain user feedback to iteratively refine system returned results, often by asking questions in visual form.
In particular, there has been a recent focus of this type of iterative refinement using relative attribute feedback~\cite{KovashkaICCV2013,KovashkaIJCV2015,ladECCV2014,modiBMVC2017}.  As seen in Figure~\ref{fig:feedback}(a), this enables a system to provide targeted feedback, but require relative attribute annotations not typically found in image datasets.




In this paper, we propose an interactive image search system which doesn't use relative attribute annotations.  Instead, we learn an image embedding trained on meta-data labels which are collected natively in e-commerce datasets.  These labels identify attributes with clear definitions (\eg does the shirt have long sleeves?), and are also be useful for other tasks such as organizing and filtering a dataset.  Relative attributes, by comparison, can be subjective in nature, and annotators may prefer to label as many as 40\% of image pairs as having equal amounts of an attribute~\cite{yuCVPR2014}, making their usefulness beyond active image search unclear.  This suggests a trade-off between the low annotation cost of our approach (due to using existing annotations) and expected performance gains from targeted feedback using relative attributes.  This paper takes a step towards characterizing the nature of this trade-off. 

We begin an interactive image search session by presenting a user with an initial set of candidate images after receiving an initial query.  A user simply selects the image which is the most visually similar to their target image (see Figure~\ref{fig:feedback}(b) for an example of the expected feedback).  We incorporate the new information provided by the user into our model and then select the next set of images. Thus, the goal is to select the most informative images to present to the user on each iteration. While this style of feedback provides less information than many relative attribute approaches, since each user determines their own similarity criteria resulting in different responses, our experiments show it is ``good enough'' in many cases.  

\begin{figure}[t]
\centering
\includegraphics[width=0.35\textwidth,trim=0cm 6.6cm 0cm 0cm, clip]{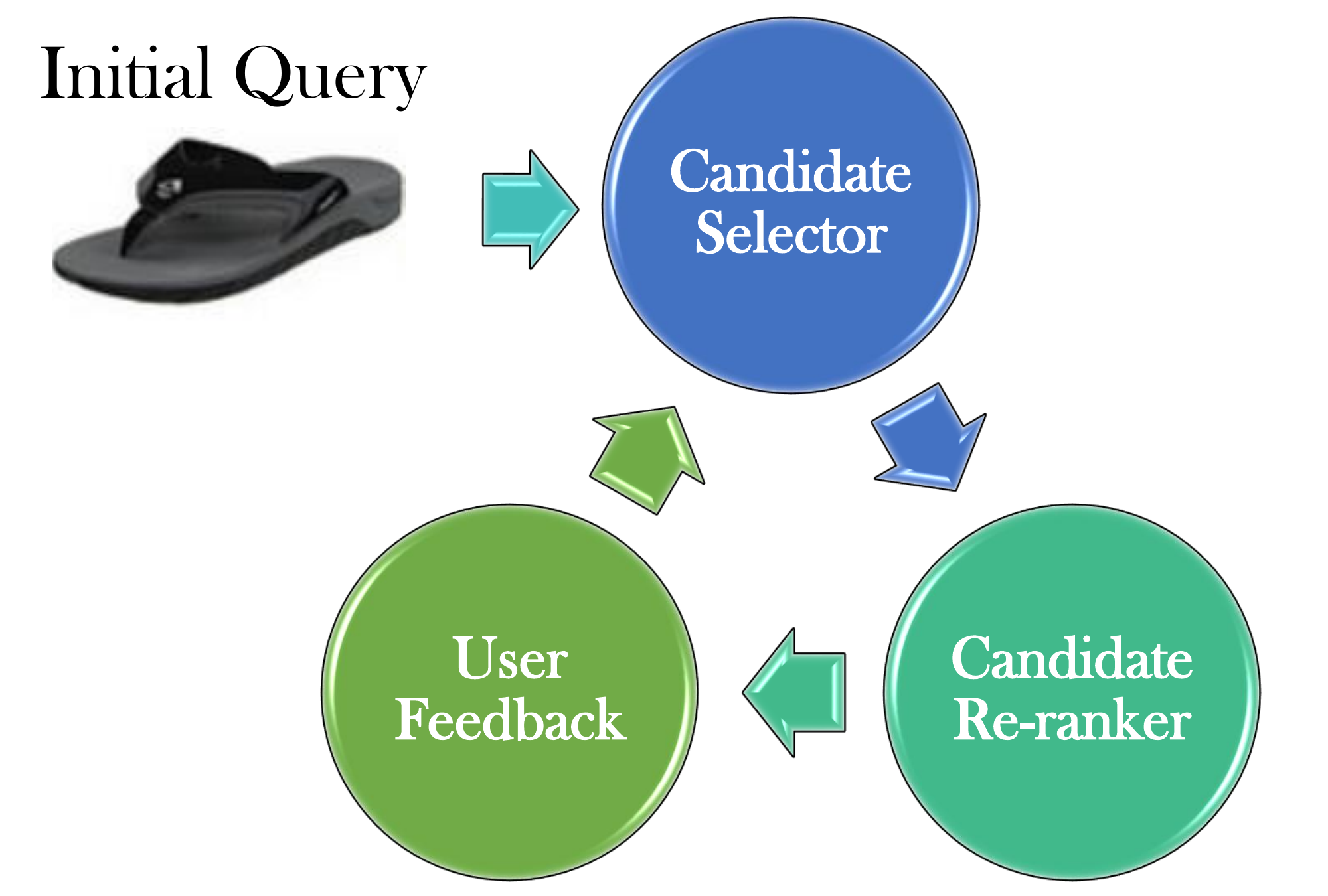}
\caption{{\bf Search refinement process.} At test time we are given an initial query as input to our system.  On each iteration, we search our database using our ``candidate selector'' strategies to obtain an initial set of candidates.  We use a ``candidate re-ranker'' on this set of images using informative, but computationally expensive selection criteria.  During the ``user feedback'' step the user indicates if the new refined candidates are more representative of their desired image or not. If they accept a new image, it becomes the query for the next iteration.}
\label{fig:query_refinement_process}
\end{figure}

A popular selection criterion is Expected Error Reduction (EER)~\cite{bransonECCV2010,KovashkaICCV2013,KovashkaICCV2011,MacAodhaCVPR2014,MensinkCVPR2011,royICML2001}.  This strategy chooses images that provide the largest reduction in the generalization error of the current model, but its high computational cost is disqualifying for many tasks.  As such, EER is typically computed on a short list of candidates (\eg exemplars from hierarchical clusters~\cite{KovashkaICCV2013,MacAodhaCVPR2014}). We experiment with two low-cost sampling strategies to obtain a candidate list in this work: a nearest neighbor baseline, which largely ignores user feedback, and a criterion that greedily selects images reflecting the feedback from prior iterations. Figure~\ref{fig:query_refinement_process} contains an overview of our active image search process.

A limitation of hand-crafted criteria like EER is the inability take advantage of the interplay between attributes.  For example, knowing the target shirt has a collar also provides some information about the type of shirt being searched for. A good sampling criteria should be able to take advantage of such information as well as adapt to user behavior.  Thus, we employ reinforcement learning using a Deep Q-Network~\cite{mnih-atari-2013} to learn to select informative images which we use to refine our list of candidates.

\begin{figure}[t]
\centering
\includegraphics[width=0.48\textwidth,trim=0 2.5cm 13cm 0,clip]{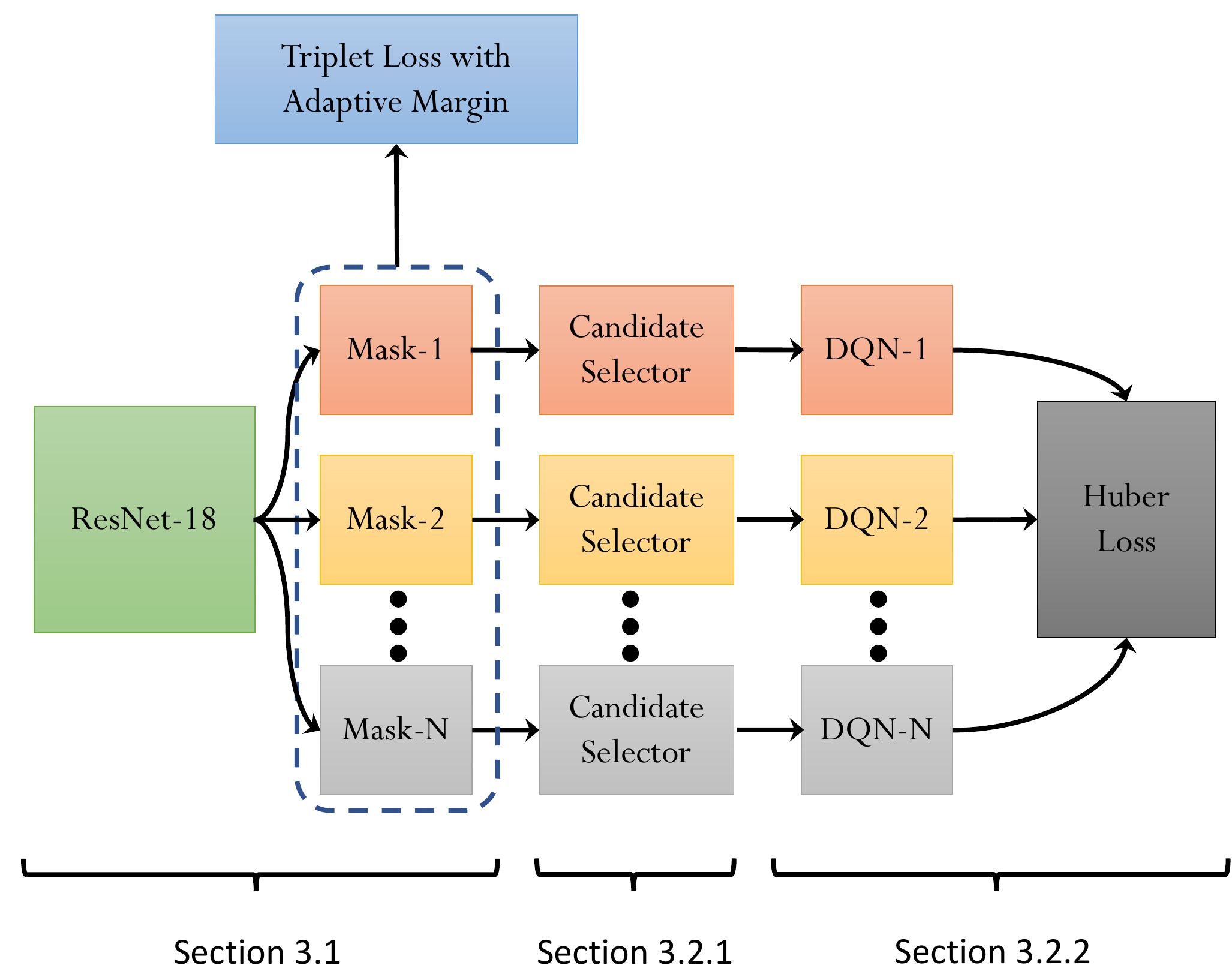}
\caption{\textbf{Model Overview.} Our model consists of three major components.  First, we train a feature extractor which computes an embedding representation for each image in our database that can be projected into an attribute specific space using a learned mask.  These are fed into our Candidate Selectors, which obtains a list of likely candidates.  Finally, from these candidates we select the most informative image according to each attribute using a DQN consisting of three fully connected layers followed by a ReLU.}
\label{fig:dqn_structure}
\end{figure}

To learn a good feature representation for our task, we introduce several enhancements to the state-of-the-art Conditional Similarity Network (CSN)~\cite{veitCVPR2017}.  This model trains a single network to learn an embedding for multiple attributes.  It accomplishes this by learning a masking function for each attribute which selects features in a general representation for an image that is important to separating images in that concept space.  This provides multiple views of the images in our database which has proven useful on similar tasks (\eg\cite{heICDM2016,KovashkaICCV2013,KovashkaIJCV2015,modiBMVC2017}) and tends to perform better than training separate embedding models for each concept. Whereas the authors of the CSN model considered the label for each embedding in isolation (\ie an embedding trained for colors would only care if both images were blue), we also factor the overall similarity between two images when training our representation.  
The resulting model encourages samples to separate into homogeneous subgroups in each embedding space. Therefore, as we traverse an attribute embedding, \eg heel height, a transition from boot to a stiletto in a single step would be unlikely even if they both have the same sized heel.  Combined with constraints which enable us to better exploit our training data, we show significant performance improvements in measuring the similarity between two images with regards to a specific concept.  We provide an overview of our approach's parts in Figure~\ref{fig:dqn_structure}.

Our primary contributions are summarized below:
\begin{itemize}
\item We build a system which refines image search results without using the relative attribute annotations or attribute inputs required in prior work.
\item We introduce enhancements to the Conditional Similarity Network which encourages smooth transitions as we traverse the learned embedding space (Section~\ref{sec:embedding}).
\item We propose a Deep-Q Network-based selection criteria instead of hand-crafted methods (Section~\ref{sec:im_sampling_strat}).
\end{itemize}

Our experiments in Section~\ref{sec:experiments} show our image representation reduces attribute matching errors by 2.5-3\% on the UT Zap-50K~\cite{yuCVPR2014} and OSR~\cite{oliviaIJCV2001} datasets while also finding a specific image in a database faster than hand-crafted sampling strategies.

\section{Related Work}
\label{sec:related}

\noindent{\bf Attribute-based interactive image search.} A key difference between this paper and prior work (\eg~\cite{KovashkaIJCV2015}) is that we train our models using annotations which already exist in many datasets.  Much of the recent work on this task has focused on how to best utilize models trained using expensive relative attribute annotations (\eg\cite{KovashkaICCV2013,KovashkaIJCV2015,modiBMVC2017}) or requires a user to specify which attributes their target image has (\eg\cite{ZhangACMM2012,Zhao_2017_CVPR}).  These assumptions provide attribute specific feedback so the model knows exactly how to alter the current image to make it more like the target image, but are more costly than our approach in terms of annotations, user requirements, or both.
\smallskip

\noindent{\bf Reinforcement learning and active learning.} Recently there has been an ever-growing trend of abandoning hand-crafted approaches in favor of learned models.  Training models for selecting informative examples in active learning, however, has primarily focused on how best to combine hand-crafted sampling strategies (\eg\cite{BaramJMLR2004,OsugiICDM2005}).  In~\cite{ebert12cvpr}, the authors used reinforcement learning to select which hand-crafted strategy to use on each iteration.  This idea was extended in~\cite{Long_2015_ICCV} to select which annotator to use as well as finding informative samples.  In contrast, our approach creates an entirely new criterion rather than combining hand-crafted strategies, sharing a similar spirit to some early work in relevance feedback (\eg~\cite{yinICCV2003,Yin:2005:IRF:1083822.1083983}).
\smallskip

\noindent{\bf Relative Attributes.}  Prior work in predicting relative attributes include using pairwise supervision to learn linear rankers~\cite{parikhICCV2011}, multi-task learning~\cite{Chen_2014_CVPR}, and fusing binary and relative attribute labels in a model that would make predictions for both types of labels~\cite{Wang_2016_CVPR}.  Some works found training local rankers could lead to improved performance~\cite{yuCVPR2014,Yu_2015_ICCV}.  Deep networks have also been used to rank attributes~\cite{souriACCV2016} as well as localize them~\cite{singhECCV2016}.  However, most of these approaches rely solely on expensive pairwise supervision in order to train their models. An exception is Yu~\etal~\cite{Yu_2017_ICCV} which augmented their annotated pairs with synthetically generated images.  In our work, we use labels which are typically found natively in many datasets rather than rely on relative attribute annotations.

%% file: attribute_embeddings.tex
\section{Image Search with Active Feedback}

Our objective is to quickly locate a target image $I_t$ in a database given a query $q$.  While the initial query can take multiple forms (\eg keywords, images~\cite{chumICCV2007,gongCVPR2011}, or sketches~\cite{yu2016sketch}), we will assume it is provided as an image $I_{q_0}$ which shares some desirable attribute with the target image.  In order to locate the target image by obtaining feedback from the user, we need a representation where we can measure similarity between images in the database as well as a selection strategy which uses this representation to find informative images to present to the user.  In this paper we provide enhancements for both learning the image representation, which we shall discuss in Section~\ref{sec:embedding}, and sampling strategies, which we will present in Section~\ref{sec:im_sampling_strat}.

\subsection{Globally-Consistent Attribute Embeddings}
\label{sec:embedding}

To compare two images, we train a set of embeddings, each representing a different attribute we wish to capture.  This provides multiple senses of each image which we can use to select informative images to the user.  Due to its state-of-the-art performance and efficiency, we chose to build upon the CSN model~\cite{veitCVPR2017} which we shall briefly review before describing our modifications.

\subsubsection{Conditional Similarity Network}
The CSN model was designed to learn a disentangled embedding for different attributes in a single model.  A general image representation is learned through the image encoding layers of the model.  Then a trained mask is applied to the representation to isolate the features important to that specific attribute.  This enables each embedding to share some common parameters across concepts, while the mask is tasked with transforming the features into a discriminative representation.  After obtaining the general embedding features between two images $G_i, G_j$, they are compared using a masked distance function,

\begin{equation}
D_{m}(G_i, G_j; m_a) = ||G_i \star m_a - G_j \star m_a ||_2,
\label{eq:mask_distance}
\end{equation}

\noindent where $m_a$ is the mask for some attribute and $\star$ denotes an element-wise multiplication. Then, given a triplet of embedding features $(G_x, G_y, G_z)$ where the pair $(G_x, G_y)$ share the same attribute label which is also not shared by $G_z$, the CSN model is trained using the margin based loss function given by

\begin{eqnarray}
\label{eq:triplet_loss}
\lefteqn{L_T(G_x, G_y, G_z; m_a) = } \\ 
& & \max\{0, h + D_{m}(G_x, G_y; m_a) - D_{m}(G_x, G_z; m_a)\} \,,
\nonumber
\end{eqnarray}

\noindent where $h$ controls the minimum margin between positive and negative pairs. The general unmasked embedding representation $G$ is L2 regularized to encourage regularly in the latent embedding space.  The masks $m$ are $L1$ regularized to encourage a sparse feature selection.  Thus, the complete loss function is

\begin{eqnarray}
\lefteqn{L_{CSN}(G_x, G_y, G_z; m_a) = } \\ \label{eq:csn_loss}
& & L_T(G_x, G_y, G_z; m_a) + \lambda_1||G||^2_2 + \lambda_2||m_a||_1 \,,
\nonumber
\end{eqnarray}

\noindent where $\lambda_{1-2}$ are scalar parameters.

We modify the original CSN model by L2 normalizing the final attribute representation (\ie $G \star m$ in Eq.~(\ref{eq:mask_distance})) as this tends to make training more stable~\cite{SchroffCVPR2015}.  In addition, since the masks can be viewed as an attention over the general embedding features we force them to sum to 1 as typically done for attention models (\eg ~\cite{xu2015show}).

\subsubsection{Incorporating Global Compatibility}
\label{sec:global_sim}
\begin{figure}
\centering
\begin{tabular}{c|cc}
Anchor Image & & \\
\includegraphics[width=0.06\textwidth]{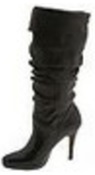} & \includegraphics[width=0.06\textwidth]{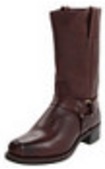} & \includegraphics[width=0.06\textwidth]{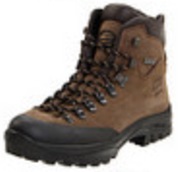}\\
\hline
Shared Attributes with Anchor: & boot &  boot\\
 & mid-calf & \\
 \hline
\end{tabular}
\caption{During training, we take into account the overall similarity between images based on the number of shared attributes. This encourages the model to maintain the left-to-right ordering of the images above in a category embedding space even though they all belong to the boot category}
\label{fig:global_sim_motivation}
\end{figure}

Since our goal is to traverse our embeddings in order to locate some target image, it is desirable that they provide natural transitions from image to image. For example, if we were to transition from the anchor image to the rightmost image in Figure~\ref{fig:global_sim_motivation} it would be considered a significant divergence.  The center image, while still different, seems like a more logical transition even though all three images belong to the boot category. Therefore, to make our embedding spaces more intuitive, we also take into account the overall similarity between two images beyond the attribute being encoded. Given the set of attributes $A_x, A_y, A_z$ for each of the images in our training triplet, we compute the difference in shared attributes between the negative and positive pairs:

\begin{equation}
w(A_x, A_y, A_z) = \max\{0, \frac{1}{\mathcal{E}}(|A_x \cap A_y| - |A_x \cap A_z|)\}
\end{equation}

\noindent where $\mathcal{E}$ represents the number of embeddings being trained.  We prevent negative values of $w$ to maintain a minimum margin between negative and positive pairs of the triplet.  We define our new margin $h'$ for Eq. (\ref{eq:triplet_loss}) as

\begin{equation}
h'(A_x, A_y, A_z) = \zeta + \eta w(A_x, A_y, A_z),
\label{eq:global_similarity}
\end{equation}

\noindent where $\eta, \zeta$ are a scalar parameters.

%% file: active_image_search.tex
\subsection{Image Sampling Strategies}
\label{sec:im_sampling_strat}

Using the representation for the images in our database from Section~\ref{sec:embedding}, our task is to select the $K$ most informative images to present to the user for feedback in order to quickly locate target image $I_t$.  We begin by obtaining a short list of candidate images in Section~\ref{sec:candidates}, before refining this list using more powerful, but computationally expensive methods in Section~\ref{sec:sampling}.

\subsubsection{Candidate Selection Methods}
\label{sec:candidates}

Most selection strategies focus on trying to reduce uncertainty in the current model, or exploit the information obtained in order to make fine-grained distinctions.  In practice, however, many search engines provide means to filter results based on meta-data labels. For example, when searching for clothing a search engine may allow you to filter results based on its category (\eg pants), subcategory (\eg jeans), and color, amongst others.  Coupled with the initial query, this provides a strong signal to initialize an active learning algorithm.  Thus, the criteria that follow focus on the exploitation of existing knowledge.
\smallskip

\noindent{\bf Nearest Neighbors.} As a baseline, we perform an iterative nearest neighbors query to obtain candidate images.  Given query image $I_q$, this method returns the $K$-nearest neighbors to $I_q$ that have not been previously selected.  Which ever image is selected as most relevant to the target image is used as the query in the next iteration.
\smallskip

\noindent{\bf Feedback Constraint Satisfaction.} Inspired by~\cite{KovashkaIJCV2015}, we find the samples which satisfy the maximum number of feedback constraints provided by the user.  For each iteration that a new candidate query $I^*_{q_{i+1}}$ is accepted by the user, then we know that $I^*_{q_{i+1}}$ is closer to the target image than $I_{q_{i}}$.  Analogously, if the candidate is not accepted, then we know $I^*_{q_{i+1}}$ is farther away from the target image than $I_{q_{i}}$. These become constraints $F$ where each element is a tuple $(I_x, I_y)$ where $I_x$ is closer to the target image than $I_y$.  We define $l$ as a binary variable which indicates that we want to count the number of unsatisfied constraints (\ie for this criterion $l = 0$ so we count satisfied constraints). Then we can calculate the portion of constraints in $F$ a candidate image $I_o$ satisfies, \ie,
\begin{equation}
S(I_o | l, F) = \frac{1}{|F|} \sum_{\forall I_{x_n}, I_{y_n} \in F} \mathbbm{1}_{fcs}(I_o, I_{x_n}, I_{y_n}) \oplus l,
\label{eq:constraint_satisfaction}
\end{equation}

\noindent where $\mathbbm{1}_{fcs}$ is an indicator function which returns one if $D(I_o, I_{x_i}) < D(I_o, I_{y_i})$ under some distance function $D$.  Thus, our criteria for the next proposed query from the set of candidates $\mathcal{O}$ is:

\begin{equation}
I^*_{q_{i+1}} = \argmax_{I_o \in \mathcal{O}} S(I_o | l = 0, F).
\end{equation}

\noindent Ties are broken using nearest neighbors sampling between the candidates and the query image.

\subsubsection{Candidate Re-ranking Methods}
\label{sec:sampling}

Many methods that measure how informative a sample is are computationally expensive, making it infeasible to run over a large database.  Therefore,  we begin by obtaining a short list of candidates using the methods from Section~\ref{sec:candidates}, then re-rank them based on how informative they are.  Below we discuss two such re-ranking methods.
\smallskip

\noindent{\bf Expected Error Reduction.}  Initially proposed in~\cite{royICML2001}, this refinement strategy focuses on reducing generalization error of the current model of the desired target image. Ergo, it can be seen as inherently balancing exploration and exploitation criteria.  We measure the entropy of the current model by calculating the portion of constraints an image satisfies as done in Eq.\ (\ref{eq:constraint_satisfaction}), \ie,

\begin{equation}
H(F) = -\sum_{I_o \in \mathcal{O}}\sum_l \hspace{2mm} S(I_o | l, F)\log(S(I_o | l, F)).
\end{equation}

\noindent  We use the highest ranked item which hasn't been presented to the user, which we denote as $I_{t^*}$, as a proxy for the target image when predicting the user's response $r$.  The simulated response either accepts or rejects some image $I_c$ from our short list of candidate images $\mathcal{C}$ to create a new constraint.  For example, $(I_q, I_c, r=0)$ would indicate that a constraint should be added to $F$ that says $I_c$ is farther away from the target image than the current query $I_q$. We decide if a new constraint would be satisfied by measuring the likelihood that the candidate image shares the same attributes with the target image.  The candidate images are then selected according to the following:

\begin{equation}
I^*_{q_{i+1}} = \argmin_{I_c \in \mathcal{C}} \sum_r \sigma(r | D(I_c, I_{t^*}), \phi) H(F \cup (I_q, I_c, r)),
\end{equation}

\noindent where $\sigma$ converts the distances in the attribute's embedding space to probabilities using Platt scaling~\cite{platt1999} whose parameters $\phi$ we estimate using the training set.  Effectively, we select the $I_c$ which we are the most uncertain about that is also similar to our best guess at the target image.
\smallskip

\noindent{\bf Learned Re-ranking Criteria.}  So far only hand-crafted strategies have been discussed.  Learned criteria can easily adapt to the exact task and dataset, making it an attractive option.  To this end, we train a Deep Q-Network (DQN)~\cite{mnih-atari-2013} with experience replay to learn how to select informative images.  In this paradigm, we learn a function $Q$ that estimates the reward $\rho$ we would get by taking some action given the current state of the system $\Psi$. We define $\rho$ as the change in the percentile rank of the target image under the current model after obtaining feedback from the user. We represent each image in the list of candidates obtained from the methods in Section~\ref{sec:candidates} as the difference between its visual embedding and the query image. This is fed into our DQN as the current state $\Psi$, which then selects which image to present to the user (\ie the set of actions asks which image to choose).  At test time, the selection criteria simply need to maximize the expected reward if we were to select image $I_c$ to present to the user:

\begin{equation}
I^*_{q_{i+1}} = \argmax_{I_c \in \mathcal{C}} Q(I_c, \Psi).
\end{equation}

\noindent Our model is trained using a Huber loss on top of the temporal difference error between expected and observed rewards.

%% file: experiments.tex
\section{Experiments}
\label{sec:experiments}

We begin by validating our image representation's ability to identify if two images share the same attributes in Section~\ref{sec:attributes}.  Then we analyze the ability of our approach to perform our active image search task in Section~\ref{sec:active_search}.
\smallskip

\noindent{\bf Dataset.} Experiments were performed on the UT Zappos50K (UT Zap-50K) dataset~\cite{yuCVPR2014}.  This dataset consists of just over 50K images taken from the Zappos website of shoes in a canonical view and homogeneous backgrounds.  Each image has eight meta-data attributes associated with it: category, closure, gender, heel, insole, material, subcategory, and toestyle.  Only the category and subcategory labels are required, resulting in a sparse labeling of the remaining attributes.  We split images in the dataset by their productID, keeping 5000 products for testing, 1000 for validation, and used the remaining for the training.

\subsection{Attribute Embedding Experiments}
\label{sec:attributes}
\noindent{\bf Implementation Details.}  We generally follow the training procedure described in~\cite{veitCVPR2017}.  For each attribute in the dataset, we randomly sampled 200K triplets for training, 40K for testing, and 20K for the validation set from their respective images.  We did not use the same triplets as~\cite{veitCVPR2017}, however, since they split their images randomly which could result in same product appearing in both the training and testing splits.  Although we tried semi-hard negative sampling of triplets~\cite{SchroffCVPR2015}, it did not provide performance benefits in our experiments.  The models were trained for 200 epochs with a batch size of 256.  The best model is selected using the validation set. We set our parameters as the following: ($h = 0.3$, $\lambda_1 = 5e^{-4}$, and $\lambda = 5e^{-6}$ in Eq. (\ref{eq:triplet_loss}), and $\eta = h$ in Eq. (\ref{eq:global_similarity}).  We initialize our model using an 18 layer Deep Residual Network~\cite{He_2016_CVPR} that was trained on ImageNet~\cite{deng2009imagenet}.  All images are resized to $112 \times 112$ before being fed into the network.

\noindent{\bf Evaluation Metric.}  Following~\cite{veitCVPR2017} we report the triplet satisfaction rate of our model (\ie the percentage of valid triplets for the 320K samples in the test set).
\smallskip

\noindent{\bf Results.}  Table~\ref{tab:csn_results} reports our triplet satisfaction rate on the test set and compares our approach to the state-of-the-art.  The first two lines of Table~\ref{tab:csn_results}(b) show that doubling the number of training triplets for the baseline model results in a very small improvement to performance.  However, the third line of Table~\ref{tab:csn_results}(b) demonstrates that by including the $L2$ normalization on the embedding outputs and forcing each mask to sum to 1 (referred to in the table as ``constraints''), we can better leverage the additional training data, improving our performance by almost 2\%.  Our full model, which includes these constraints as well as our attention to the global similarity between images (described in Section~\ref{sec:global_sim}) results in a 3\% improvement over the baseline.

\begin{table}[t]
  \setlength{\tabcolsep}{3pt}
  \centering
    \begin{tabular}{|l|c|c|}
      \hline
      \multirow{ 2}{*}{Method} & triplets/ & \multirow{ 2}{*}{Accuracy}\\ 
      & concept & \\
      \hline
      \hline
      CSN & 50K & 79.29\\
      CSN & 100K & 79.35\\
      CSN + constraints & 100K & 81.27\\
      CSN + constraints + global similarity & 100K & \textbf{82.28}\\
      \hline
    \end{tabular}
    \caption{Triplet satisfaction rate and number of training triplets used for the UT-Zap50K dataset~\cite{yuCVPR2014}. Note: we used all eight meta-data labels rather than just the four reported in Veit~\etal~\cite{veitCVPR2017}.}
\label{tab:csn_results}
\end{table}

\begin{table*}[t]
\setlength{\tabcolsep}{4.8pt}
  \centering
    \begin{tabular}{|l|c|c|c|c|c|c|c|c|}
      \hline
      Method & Category & Closure & Gender & Heel & Insole & Material & Subcategory & Toestyle\\
      \hline
      \hline
      CSN & 93.69 & 77.17 & 77.27 & 88.49 & 58.64 & 71.53 & 90.21 & 77.31\\
      CSN + constraints & 93.07 & 79.80 & 80.15 & 89.40 & 60.35 & 75.00 & 92.90 & 79.46\\
      CSN + constraints + global similarity & \textbf{94.48} & \textbf{81.63} & \textbf{81.37} & \textbf{89.62} & \textbf{61.68} & \textbf{75.75} & \textbf{93.98} & \textbf{79.83}\\
      \hline
    \end{tabular}
    \caption{Triplet satisfaction rate using models trained with 100K triplets/concept on the UT-Zap50K dataset~\cite{yuCVPR2014} separated by attribute.}
\label{tab:csn_concept_results}
\end{table*}

\begin{figure}[t]
\centering
\includegraphics[width=0.48\textwidth,trim=.9cm 0cm 1.6cm 0cm,clip]{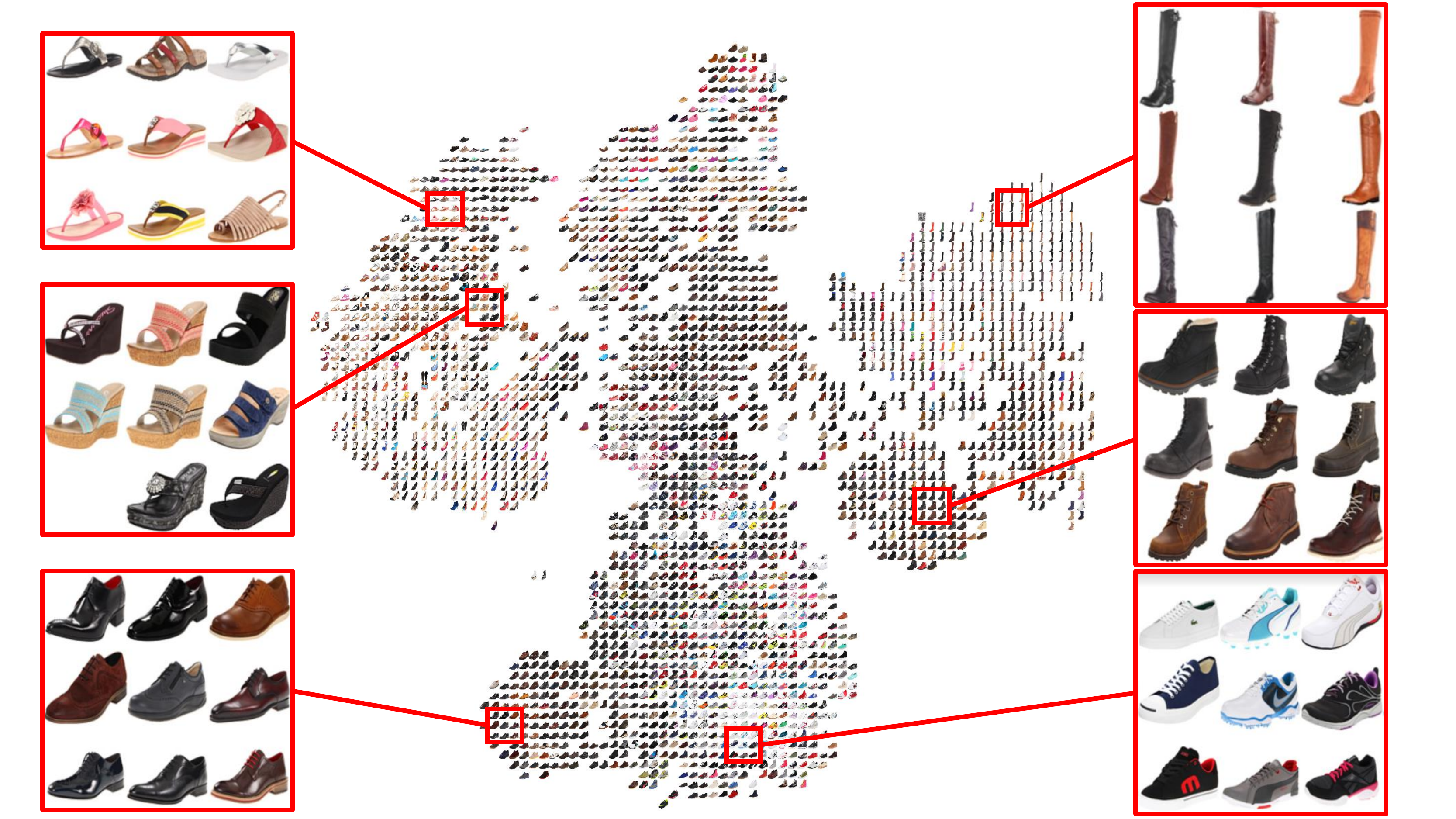}
\caption{{\bf Visualization of the learned embedding.} t-SNE~\cite{vandermatten2008JMLR} of the closure embedding space using our improved CSN Model.  Starting from the top left and moving down and to the right, in the first two pair of boxes we see that the embedding has learned to separate sandals based on heel-height despite both being slip-ons.  The next three show that the embedding has learned subcategories of shoes despite all having lace-up closure mechanisms, with the last pair showing how different closing mechanisms for boots have also been separated.}
\label{fig:tsne}
\end{figure}

We break down the performance of our model by the attribute being learned in Table~\ref{tab:csn_concept_results}.  The material and gender attributes reported the largest performance improvement at 4\% over the baseline CSN model.  The subcategory attribute brought up its performance by just over 3.5\%, putting its performance more in line with the category attribute.  While our model did improve the category attribute by almost 1\%, including just the constraints did lower performance slightly.  However, this loss was more than made up by the improvements in the other attributes, and the model which included global similarity did best across all attributes.

To provide insight into the structure of the learned embedding spaces, we provide a t-SNE visualization~\cite{vandermatten2008JMLR} of the closure attribute in Figure~\ref{fig:tsne}\footnote{The closure attribute embedding is also provided in Figure 5(a) of Veit~\etal~\cite{veitCVPR2017} which mixes sandals, heels, and slippers in the same local space when not encouraging globally consistent embeddings.}.  The highlighted boxes show how our embedding has learned to separate shoes with heels from those without, or athletic and dress shoes, despite having the same closing mechanism.  This demonstrates how our representation can make more intuitive transitions while navigating the learned embedding space.

\subsection{Active Image Search Experiments}
\label{sec:active_search}

\begin{figure*}[t]
\centering
\includegraphics[width=0.73\textwidth,trim=0cm 0.5cm 5.5cm 0cm,clip]{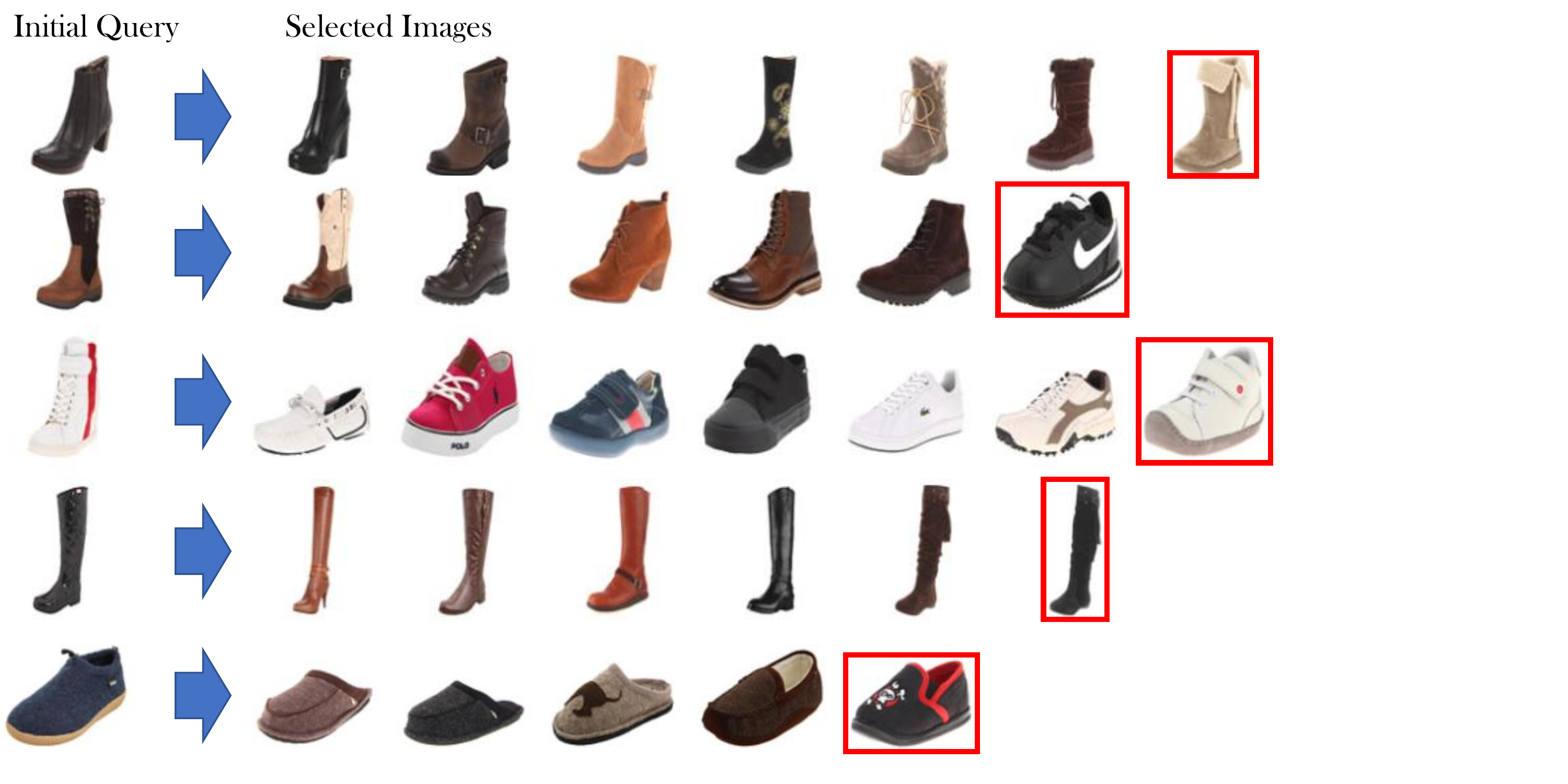}
\caption{{\bf Active Image Search Examples.} Each row is an example of the images selected by our system as we refine our search results going from our initial query to our target image which is contained by a red box}
\label{fig:traversal_example}
\end{figure*}

\begin{figure*}[t]
\centering
\topinset{\bfseries(a)}{\topinset{\bfseries(b)}{\includegraphics[width=0.95\textwidth,trim=.2cm 0.8cm .2cm 0.8cm,clip]{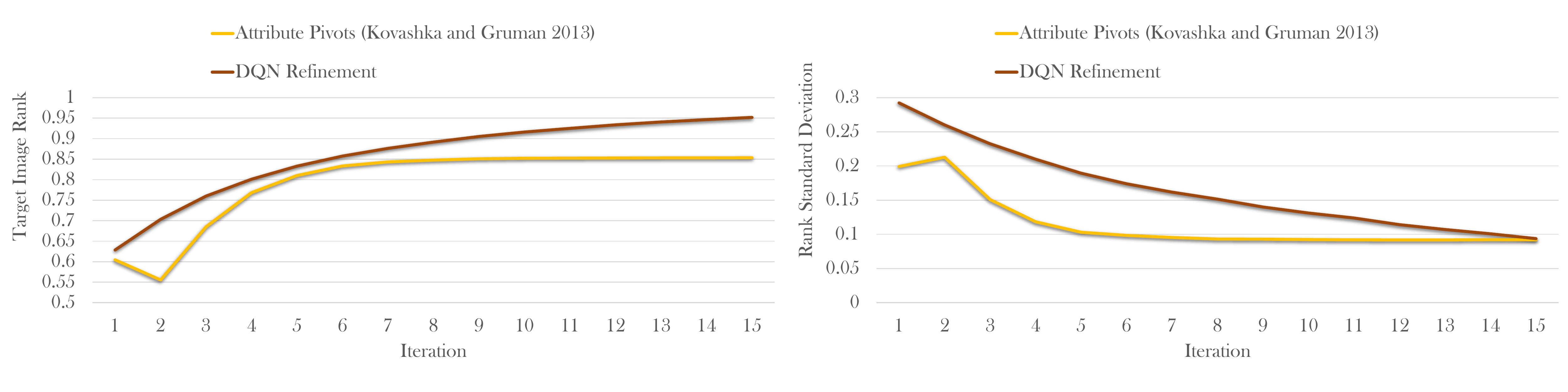}}{-.0in}{.25in}}{0.0in}{-3.05in}
\caption{Comparison of our DQN Refinement strategy over an embedding trained using binary attribute labels and the Attribute Pivots method~\cite{KovashkaICCV2013} which utilizes relative attribute annotations.}
\label{fig:target_ranking}
\end{figure*}

\noindent{\bf Implementation Details.} On each iteration, we select one image per attribute type (8 total) to present to the user.  For our refinement experiments, we select 4 candidates per attribute using our Feedback Constraint Satisfaction criterion (32 total) and then re-rank them to select the top 8 images. In our DQN experiments, we use a replay memory size of 20,000, a discount rate of 0.999, and a batch size of 2048.  Our DQN models are trained using simulated user feedback which we describe below.  During training, we begin by performing a random action 90\% of the time and decay this randomized action rate exponentially until we reach 5\%. 
\smallskip

\noindent{\bf Query-Target Pairs.} For each attribute in the dataset, we sample 2,000 pairs for training, 500 for validation, and 1,000 for testing resulting in 16,000, 4,000, and 8,000 pairs, respectively.  
Each sample was randomly selected from the set of image pairs which share at least one attribute without restrictions.  This means pairs can be semantically distant from each other (\eg a boot and a sandal without a heel can be sampled for pairs sharing that attribute), adding to the challenges faced by our model.
\smallskip

\noindent{\bf User Feedback.} User feedback is simulated by averaging the Euclidean distance between the candidate image and the target image between all embedding spaces and selecting the closest one.  We evaluated the appropriateness of our user feedback mechanism by presenting triplets of images to 13 human annotators.  Each triplet contained a target image and a pair of images to evaluate.  Annotators were asked to select the image from the pair which most resembles the target image.  We obtained 100 triplets selected at random from our active learning experiments (\ie they exactly reflected the decision process made by our algorithm).  After removing triplets where at least 5 annotators disagreed with the majority, our simulated input agreed with the human annotators 79\% of the time over the remaining 86 samples. Human performance was similar, as individual annotators agreed with the majority 74-88\% of the time (84\% average).  We also experimented with two other embeddings to use for our simulated feedback: features from our fine-tuned ResNet-18 model initially trained on ImageNet and the unmasked embedding representation output by our model.  Relative performance remained consistent across feedback types in our active learning experiments. 
\smallskip

\noindent{\bf Comparison to Relative Attribute Approaches.} In addition to our own baselines, we also adapt our embedding approach to produce relative attribute scores for four common concepts using the annotations provided in~\cite{yuCVPR2014}.  We encode a pair of images $I_i, I_j$ using the Conditional Similarity Network to obtain an embedding representation for an attribute.   We concatenate together the embedding from each image which is fed into a fully connected layer followed by a softmax with an output dimension of 3 and is trained jointly with the embeddings.  The output of this model indicates if image $I_i$ has more, less, or the same amount of an attribute as image $I_j$.  Using this model, we reproduce the binary tree EER-based approach of~\cite{KovashkaICCV2013}.   We initialize the activate image search model using the query image provided at test time.  The remaining implementation details follow~\cite{KovashkaICCV2013}.  We shall refer to this reproduction as Attribute Pivots henceforth.
\smallskip

\noindent{\bf Evaluation Metric.} We evaluate performance based on how many iterations were required to go from the initial query to the target image.
\smallskip

\noindent{\bf Results.}  We report our active image search performance in Table~\ref{tab:query_refinement_performance_multispace}\footnote{We don't include a comparison to Attribute Pivots~\cite{KovashkaICCV2013} in Table~\ref{tab:query_refinement_performance_multispace} as it regularly satisfied its stopping criterion (\ie~it was no longer able to improve its model) before finding the target image.  Altering the stopping criterion so that it would only stop when it located the target image resulted in poor performance on this task.}.  As seen in the top two lines of the table, using the feedback constraint satisfaction criterion reduces the number of iterations required to find a target image by 2 over the nearest neighbors baseline. Refining the top 4 candidates using expected error reduction reduces the number of steps by 1.5, and our DQN refinement reduces this further, making the total reduction approach 5 iterations fewer than the baseline.  It is important to note that this would be considered the toughest settings for this task. In practice, a user could remove a lot of images from consideration by filtering by the meta-data labels.

\begin{table}[t]
\centering
\begin{tabular}{|l|c|}
\hline
Selection Strategy & \#Steps\\
\hline
\hline
Nearest Neighbor & 26.40\\
Feedback Constraint Satisfaction & 24.62\\
Expected Error Reduction & 23.07\\
DQN Refinement & 21.79\\
\hline
\end{tabular}
\caption{Active image search performance on UT Zap-50K.}
\label{tab:query_refinement_performance_multispace}
\end{table}

We provide examples of the images selected by our system as we refine image search results in Figure~\ref{fig:traversal_example}.  In the first row, we see how the boots change in style on each iteration, deciding on the heel first before refining the style of the boot.  The second row demonstrates how our system is capable of even changing the category of the shoe, traversing from a boot to a sneaker.  The third row shows how the system switched from changing the style of shoe to the type of closing mechanism before locating the target image, with the last two rows demonstrating how our system can handle even relatively fine-grained differences between the initial query and the target image.

A good search refinement algorithm need not produce the exact target image in the refinement stage, but simply obtain enough information that the image ranks sufficiently high in the search results.  To this end, we provide the rank of the target image per iteration in Figure~\ref{fig:target_ranking}(a) with a comparison to our implementation of the Attribute Pivots approach~\cite{KovashkaICCV2013} which takes advantage of relative attribute annotations.  Here we see the two methods perform comparably despite our method using only binary attribute labels.  It is important to note that Attribute Pivots produced a more consistent algorithm as exemplified with the lower per-iteration rank standard deviation seen in Figure~\ref{fig:target_ranking}(b), even if the average performance was slightly lower than ours.  This may be due to the limited  number of relative attributes available in the UT Zap-50K dataset, which suggests it would beneficial to further explore the trade-off between annotation cost vs.\ performance in future work.


%% file: scene_experiments.tex
\subsection{OSR Experiments}
\label{sec:osr}
To demonstrate our approach's ability to generalize we provide experiments on the Outdoor Scene Recognition (OSR) dataset~\cite{oliviaIJCV2001}. This dataset consists of 2688 images with six attributes annotated for the eight scene categories.  We randomly sampled 400 images for the test set (50/category), 160 images for the validation set (20/category), and used the rest for training.  To train our representation, we randomly sampled 100K triplets for training, 40K for testing, and 20K for validation.  In our active image search experiments we sampled 16,000 pairs for training, 4,000 pairs for testing, and 800 pairs for validation.  All other settings are the same that were used for the UT Zap-50K dataset.
\smallskip

\noindent{\bf Results.} As seen in our attribute experiments in Table~\ref{tab:scene_csn_results}, our additional constraints and global similarity enhancements provide a 1.5\% and 2.5\% improvement over the baseline, respectively.  Our results on the active image search task in Table~\ref{tab:scene_query_refinement_performance_multispace} also follow the results on UT Zap-50K, where our DQN refinement strategy outperforms the EER alternative as well as the feedback constraint satisfaction and nearest neighbor baselines.  Despite the OSR dataset being from a very different domain from UT Zap-50K, our model still provides a performance improvement over prior work.

\begin{table}[t]
  \centering
    \begin{tabular}{|l|c|}
      \hline
      Method & Accuracy\\ 
      \hline
      CSN & 96.84\\
      CSN + constraints  & 98.58\\
      CSN + constraints + global similarity & \textbf{99.42}\\
      \hline
    \end{tabular}
    \caption{Triplet satisfaction rate on the OSR dataset.}
    \label{tab:scene_csn_results}
\end{table}

\begin{table}[t]
\centering
\begin{tabular}{|l|c|}
\hline
Selection Strategy & \#Steps\\
\hline
\hline
Nearest Neighbor & 6.21\\
Feedback Constraint Satisfaction & 5.57\\
Expected Error Reduction & 4.92\\
DQN Refinement & \textbf{4.54}\\
\hline
\end{tabular}
\caption{Active image search performance on the OSR dataset}
\label{tab:scene_query_refinement_performance_multispace}
\end{table}

%% file: conclusion.tex
\section{Conclusion}
In this paper, we addressed the problem of active image search, but without expensive annotations or user requirements used in prior work.  Instead, we introduced enhancements to the Conditional Similarity Network which improved its ability to make relative attribute comparisons.  We used this representation in our experiments on active image search where we demonstrated the effectiveness of our DQN selection criterion and showed it was competitive with prior work which used expensive relative attribute annotations.  In future work, we would like to build upon our current system by taking advantage of a hierarchical clustering method to organize our data which has proven useful in prior work~\cite{KovashkaICCV2013,MacAodhaCVPR2014}.  Our model could also benefit from taking into account the diversity of selected images on each iteration by incorporating elements used in batch mode active learning approaches (\eg~\cite{Elhamifar_2013_ICCV,guoNIPS2008,kulesza2012determinantal}).